\documentclass[runningheads]{llncs}
\usepackage{url}
\usepackage{cite}
\usepackage{graphicx}
\usepackage{amsmath,epsfig}
\usepackage{booktabs}
\usepackage{subfigure}
\usepackage{diagbox}
\usepackage{multirow}
\usepackage{xcolor}
\definecolor{GT}{RGB}{0, 255,0}
\definecolor{BBS}{RGB}{255,128,0}
\definecolor{CoTM}{RGB}{0,255,255}
\definecolor{DDIS}{RGB}{255,255,0}
\definecolor{DIM}{RGB}{255,0,0}
\definecolor{NCC}{RGB}{227,207,87}
\definecolor{QATM}{RGB}{128,42,42}
\definecolor{SSD}{RGB}{0,0,255}
\definecolor{ZNCC}{RGB}{153,51,250}
\usepackage{tikz}
\usepackage{pgfplots}
\begin{document}
\title{Robust Template Matching via Hierarchical Convolutional Features from a Shape Biased CNN}
\titlerunning{Robust Template Matching via Shape Biased CNN}
%
\author{Bo Gao\inst{1}\orcidID{0000-0003-3930-1815} \and
Michael W. Spratling\inst{1}\orcidID{0000-0001-9531-2813}}
\authorrunning{Gao, B. and Spratling, M.W.}
%
\institute{Department of Informatics, King’s College London, London, UK 
\email{\{bo.gao,michael.spratling\}@kcl.ac.uk}}
\maketitle              
\begin{abstract}
Finding a template in a search image is an important task underlying many computer vision applications. Recent approaches perform template matching in a deep feature-space, produced by a convolutional neural network (CNN), which is found to provide more tolerance to changes in appearance.  In this article we investigate if enhancing the CNN's encoding of shape information can produce more distinguishable features that improve the performance of template matching. This investigation results in a new template matching method that produces state-of-the-art results on a standard benchmark. To confirm these results we also create a new benchmark and show that the proposed method also outperforms existing techniques on this new dataset. 
Our code and dataset is available at: \Url{https://github.com/iminfine/Deep-DIM}.

\section{Introduction}
Template matching is a technique to find a rectangular region of an image that contains a certain object or image feature. It is widely used in many computer vision applications such as object tracking \cite{bertinetto2016fully,ma2018robust}, object detection \cite{ahuja2013object,dai2016r} and 3D reconstruction \cite{scharstein2002taxonomy,chhatkuli2014stable}. A similarity map is typically used to quantify how well a template matches each location in an image. Traditional template matching methods calculate the similarity using a range of metrics such as the normalised cross-correlation (NCC), the sum of squared differences (SSD) or the zero-mean normalised cross correlation (ZNCC) applied to pixel intensity or color values. However, because these methods 
rely on comparing the values in the template with those at corresponding locations in the image patch they are sensitive to  changes in lighting conditions, non-rigid deformations of the target object, or partial occlusions, which results in a low similarity score when one or multiple of these situations occur. 

With the help of deep features learned from convolutional neural networks (CNNs), vision tasks such as image classification \cite{chan2015pcanet,wang2017residual}, object recognition \cite{liang2015recurrent,wohlhart2015learning}, and object tracking \cite{bertinetto2016fully,ma2018robust} have recently achieved great success. 
In order to succeed in such tasks, CNNs need to build internal representations that are less effected by changes in the appearance of objects in different images. To improve the tolerance of template matching methods to changes in appearance recent methods have been successfully applied to a feature-space produced by the convolutional layers of a CNN \cite{cheng2019qatm,kat2018matching,kim2017robust,oron2017best,talmi2017template}.

The higher layers of CNNs are believed to learn representations of shapes from low-level features
\cite{kriegeskorte2015deep}. However, a recent study \cite{geirhos2018imagenet} demonstrated that ImageNet-trained CNNs are biased toward making categorisation decisions based on texture rather than shape. This work also showed that CNNs could be trained to increase sensitivity to shape and that this would improve accuracy and robustness both of object classification and detection. Assuming that shape information is also useful for template matching, these results suggest that the performance of template matching methods applied to CNN generated feature-spaces could potentially also be improved by training the CNN to be more sensitive to shape. 

In this article we verified the assumption by comparing the features from four CNN models with a same network structure but differing in shape sensitivity. Our results show that training a CNN to learn about texture while biasing it to be more sensitive to shape information, can improve template matching performance. Furthermore, by comparing template matching performance when using feature-spaces created from all possible combinations of one, two and three convolutional layers of the CNN it was found that the best results were produced by combining features from both early and late layers. Early layers of a CNN encode lower-level information such as texture, while later layers encode more abstract information such as object identity. Hence, both sets of results (the need to train the CNN to be more sensitive to shape and the need to combine information for early and late layers) suggest that a combination of texture and shape information is beneficial for template matching. 

Our main contributions are summarized as follows: (1) We created a new benchmark which, compared to the existing standard benchmark, is more challenging, provides a far larger number of images pairs, and is better able to discriminate the performance of different template matching methods. (2) By training a CNN to be more sensitive to shape information and combining features from both early and late layers, we created a feature-space in which the performance of most template matching algorithms is improved. (3) Using this feature-space together with an existing template matching method, DIM \cite{spratling2019explaining}, we obtained state-of-art results on both the standard and new datasets.
\section{Related Work}
\label{sec-related}

To overcome the limitations of classic template matching methods, many approaches  \cite{cheng2019qatm,kat2018matching,oron2017best,talmi2017template} have been developed. These methods can be classified into two main categories. 

\textbf{Matching}. One category changes the computation that is performed to compare the template to the image to increase tolerance to changes in appearance. For example, Best-Buddies Similarity (BBS) counts the proportion of sub-regions in the template and the image patch that are Nearest-Neighbor (NN) matches \cite{oron2017best}. Similarly, Deformable Diversity Similarity (DDIS) explicitly considers possible template deformation and uses the diversity of NN feature matches between a template and a potential matching region in the search image \cite{talmi2017template}. The Divisive Input Modulation (DIM) algorithm \cite{spratling2019explaining} extracts additional templates from the background and lets the templates compete with each other to match the image. 
Specifically, this competition is implemented as a form of probabilistic inference known as explaining away \cite{kersten2004object,spratling2012unsupervised} which causes each image element to only provide support for the template that is the most likely match. Previous work has demonstrated that DIM, when applied to color feature-space, is more accurate at identifying features in an image compared to both traditional and recent state-of-the-art matching methods \cite{spratling2019explaining}.

\textbf{Features}. The second category of approaches changes the feature-space in which the comparison between the template and the image is performed. The aim is that this new feature-space allows template matching to be more discriminative while also increasing tolerance to appearance changes. Co-occurrence based template matching (CoTM) transforms the points in the image and template to a new feature-spaced defined by the co-occurrence statistics to quantify the dissimilarity between a template to an image \cite{kat2018matching}. Quality-aware template matching (QATM) is a method that uses a pretrained CNN model as a feature extractor. It learns a similarity score which reflects the (soft-) repeatness of a pattern using an algorithmic CNN layer \cite{cheng2019qatm}. 

\textbf{Deep Features}. Many template matching algorithms from the first category above, can be applied to deep features as well as directly to color images. The deep features used by BBS, CoTM and QATM are extracted from two specific layers of a pre-trained VGG19 CNN \cite{simonyan2014very}, \textit{conv1\_2} and \textit{conv3\_4}. Following the suggestion in \cite{ma2018robust} for object tracking, DDIS also takes features from a deeper layer: fusing features from layers \textit{conv1\_2}, \textit{conv3\_4} and \textit{conv4\_4}. \cite{kim2017robust} proposed a scale-adaptive strategy to select a particular individual layer of a VGG19 to use as the feature-space according to the size of template. In each case using deep features was found to significantly improve template matching performance compared to using color features. 

A recent study showed that ImageNet-trained CNNs are strongly biased towards recognising textures rather than shapes \cite{geirhos2018imagenet}. This study also demonstrated that the same standard architecture (ResNet-50 \cite{ren2015faster}) that learns a texture-based representation on ImageNet is able to learn a shape-based representation when trained on ‘Stylized-ImageNet’: a version of ImageNet that replaces the texture in the original image with the style of a randomly selected painting through AdaIN style transfer \cite{huang2017arbitrary}. This new shape-sensitive model was found to be more accurate and robust in both object classification and detection tasks. Inspired by the findings, in this paper we investigate if enhancing the shape sensitivity of a CNN can produce more distinguishable features that improve the performance of template matching.
\section{Methods}
Previous work on template matching in deep feature-space (see section \ref{sec-related}) has employed a VGG19 CNN. To enable a fair comparison with those previous results, we also used the VGG19 architecture. However, we used four VGG19 models that differed in the way they were trained to encode different degrees of shape selectivity by the same approach used by  \cite{geirhos2018imagenet} (as summarised in Table \ref{tab:table1}).
\begin{table}[tb]
\caption{Four different VGG19 CNN models used in this paper. IN and SIN are the abbreviations of ImageNet and Stylized-ImageNet respectively.}
\centering
\begin{tabular}{ccccccc}
\toprule[1pt]
Name                             &Training &          Fine-tuning& Rank of shape sensitivity&\\
\hline
Model\_A                         &IN       &                    -&4 &\\
Model\_B                         &SIN      &                    -&1 &\\
Model\_C                         &IN+SIN   &                    -&2 &\\
Model\_D                         &IN+SIN   &                   IN&3 &\\
\bottomrule[1pt]
\end{tabular}
\label{tab:table1}
\end{table}
Model\_A was trained using the standard ImageNet dataset \cite{simonyan2014very} (we used the pretrained VGG19 model from the PyTorch torchvision library) which has the least shape bias. Model\_B was trained on the Stylized-ImageNet dataset and thus has the most shape bias. Model\_C was trained on a dataset containing the images from both ImageNet and Stylized-ImageNet. Model\_D was initialised with the weights of Model\_C and then fine-tuning on ImageNet for 60 epochs using a learning rate of 0.001 multiplied by 0.1 after 30 epochs. Therefore Model\_C and Model\_D have intermediate levels of shape bias, with model\_D being less selective to shape than Model\_C. The learning rate was 0.01 multiplied by 0.1 after every 30 epochs for Model\_B and after every 15 epochs for Model\_C. Number of epochs was 90 for Model\_B and 45 for Model\_C (as the dataset used to train Model\_C was twice as large as that used to train Model\_B the number of weight updates was the same for both models). The other training hyperparameters used for each model were: batch size 256, momentum 0.9, and weight decay 1e-4. The optimizer was SGD.

In color feature-space the DIM algorithm was previously found to produce the best performance (see section~\ref{sec-related}). We therefore decided to use this algorithm to determine the best CNN feature-space to use for template matching. The DIM algorithm was applied to deep features in exactly the same way that it was previously applied to color images \cite{spratling2019explaining}, except: (1) 
five (rather than four) additional templates were used; and, (2) the positive and rectified negative values produced by a layer of the CNN were directly separated into two parts and used as separate channels for the input to the DIM algorithm (in contrast, previously each channel of a color image was processed using a difference of Gaussians filter and the positive and rectified negative values produced were used as separate channels for the input to the DIM algorithm).

\section{Results}

\subsection{Dataset Preparation}

The BBS dataset \cite{oron2017best} has been extensively used for the  quantitative evaluation of template matching algorithms  \cite{cheng2019qatm,kat2018matching,kim2017robust,oron2017best,talmi2017template}. This dataset contains 105 template-image pairs which are sampled from 35 videos (3 pairs per video) from a tracking dataset \cite{wu2015object}. Each template-image pair are taken from frames of the video that are 20 frames apart. To evaluate the performance of a template matching algorithm the intersection-over-union (IoU) is calculated between the predicted bounding box and the ground truth box for the second image in the pair. The overall accuracy is then determined by calculating the area under the curve (AUC) of a success curve produced by varying the threshold of IoU that counts as success. 

\begin{figure}[bt]
  \centering
  \includegraphics[width=0.7\linewidth]{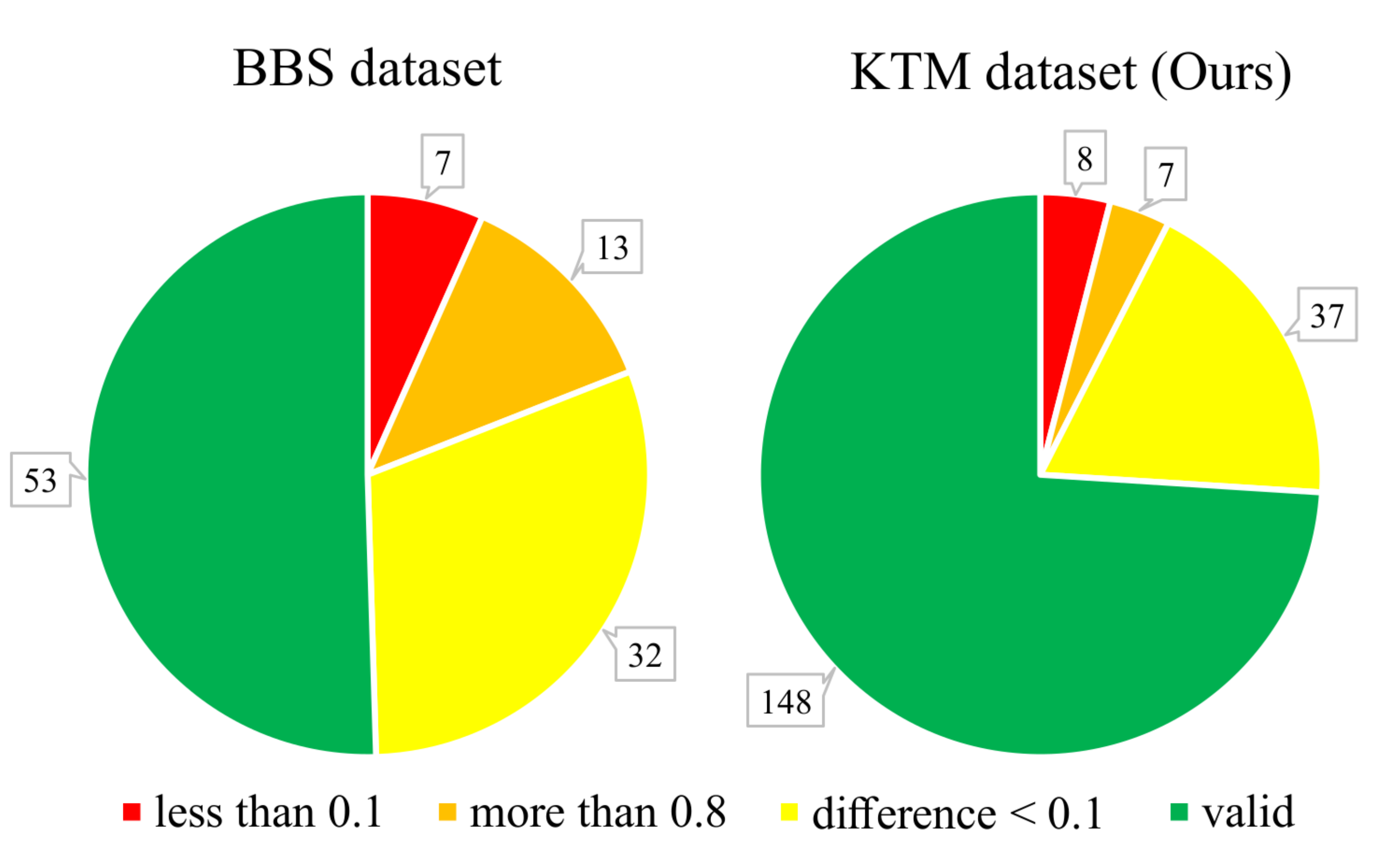}
  \caption{Discriminative ability of two datasets evaluated by comparing the IoU scores produced by ZNCC, BBS, DDIS and DIM.}
  \label{fig:fig2}
\end{figure}

Although the BBS data is widely used, it is not particularly good at discriminating the performance of different template matching methods. To illustrate this issue we applied one baseline method (ZNCC) and three state-of-art methods (BBS, DDIS and DIM) to the BBS dataset in color space. The results show that there are 52 template-image pairs where all methods generate very similar results: these can be sub-divided into 7 template-image pairs for which all methods fail to match (IoU less than 0.1 for all four methods), 13 template-image pairs for which all methods succeed (IoU greater than 0.8 for all four methods), and 32 template-image pairs for which all methods produce similar, intermediate, IoU values within 0.1 of each other. This means that only 53 template-image pairs in the BBS dataset help to discriminate the performance of these four template matching methods. These results are summarised in Figure~\ref{fig:fig2}.

We therefore created a new dataset, the King's Template Matching (KTM) dataset, following a similar procedure to that used to generate the BBS dataset. The new dataset contains 200 template-image pairs sampled from 40 new videos (5 pairs per video) selected from a different tracking dataset \cite{liang2015encoding}. In contrast to the BBS dataset, the template and the image were chosen manually to avoid pairs that contain significant occlusions and non-rigid deformations of the target that no method is likely to match successfully, and the image pairs were separated by 30 (rather than 20) frames to reduce the number of pairs for which matching would be easy for all methods. These changes make the new data more challenging and provide a far larger number of images pairs that can discriminate the performance of different methods, as shown in Figure~\ref{fig:fig2}. Both the new dataset and the BBS dataset were used in the following experiments.

\subsection{Template matching using features from individual convolutional layers}
To reveal how the shape bias effects template matching, we calculate AUC using DIM with features from every single convolutional layer of the four models. As the features from the later convolutional layers are down-sampled using max-pooling, by a factor of $\tfrac12$, $\tfrac14$, $\tfrac18$, and $\tfrac{1}{16}$ compared to the original image, the bounding box of the template is also multiplied by the same scaling factor and the resulting similarity map is resized back to the original image size to make the prediction. The AUC scores across the BBS and DIM datasets are summarised in Figure~\ref{fig:fig3}.
  
\begin{figure}[tb]
\centering
\subfigure[Evaluation on BBS dataset.]{
\includegraphics[width=0.475\linewidth]{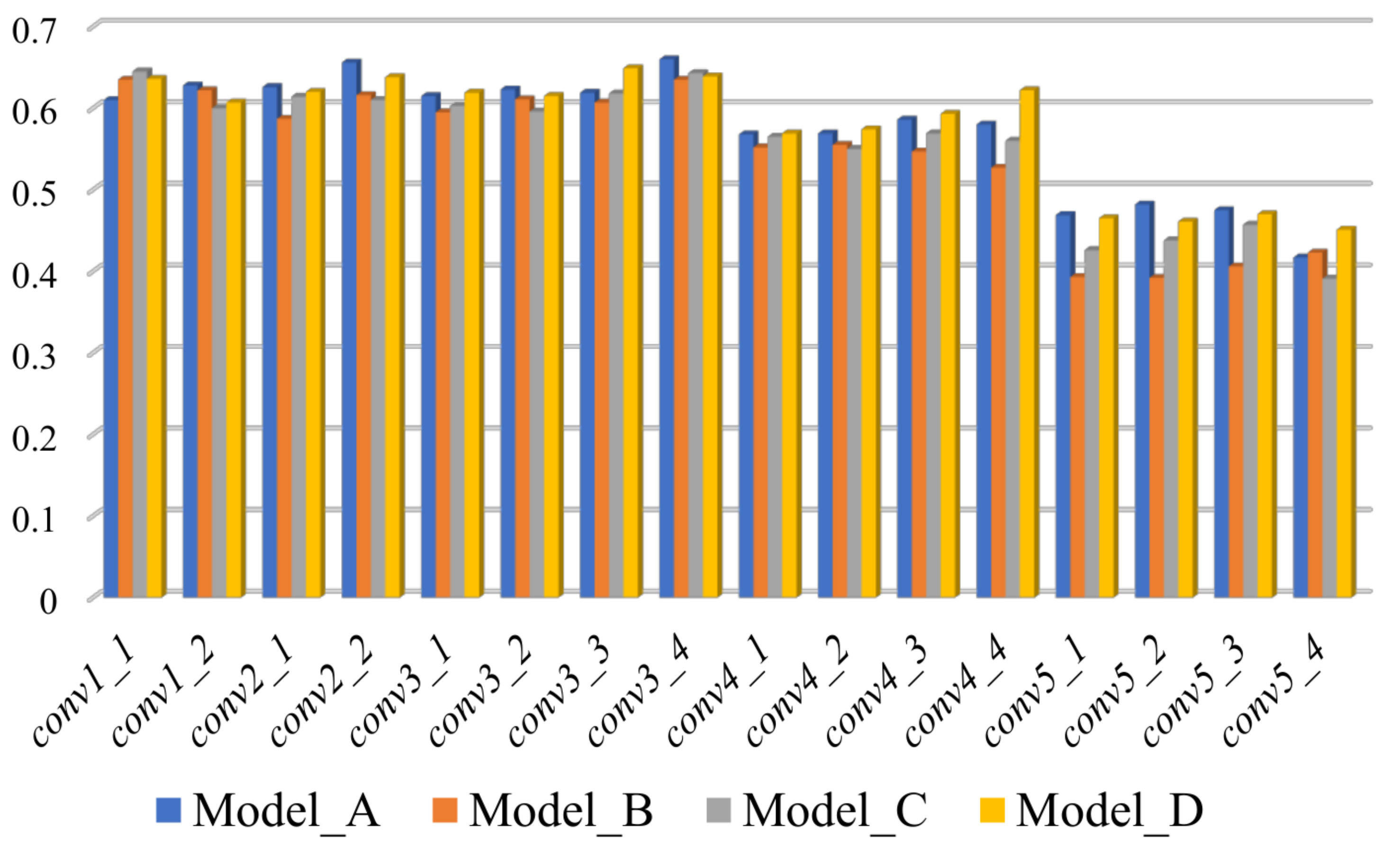}}
\subfigure[Evaluation on KTM dataset.]{
\includegraphics[width=0.475\linewidth]{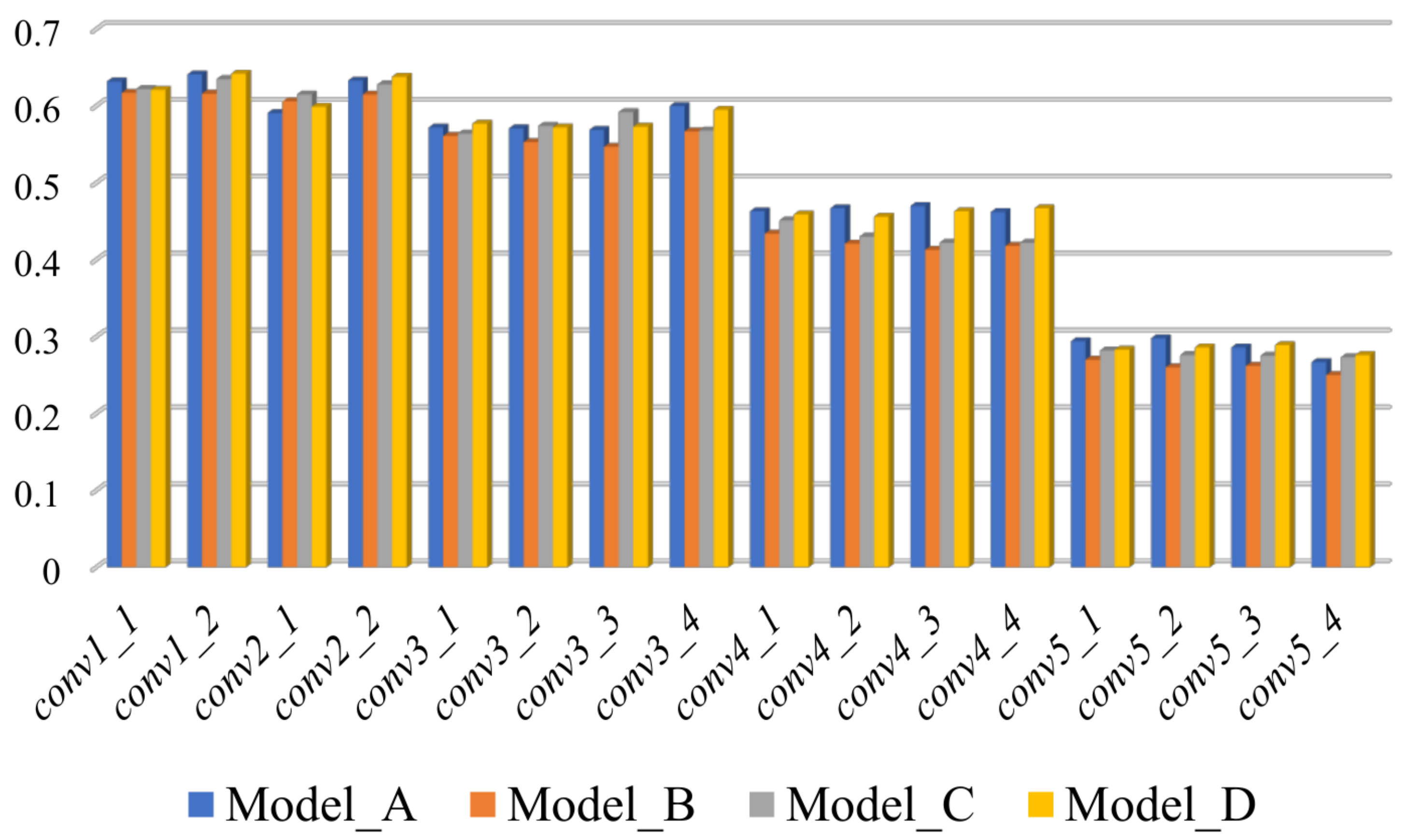}}
\caption{The AUC scores of DIM using features from different convolutional layers of four models.} 
\label{fig:fig3}
\end{figure}

For all four models there is a tendency for the AUC to be higher when template matching is performed using lower layers of the CNN compared to later layers. This suggests that template matching relies more on low-level visual attributes, such as texture, rather than higher-level ones such as shape. Among the three models trained with Stylized-ImageNet, the AUC score for most CNN layers is greater for Model\_D than Model\_C, and greater for Model\_C than Model\_B. This also suggests that template matching relies more on texture features than shape features.  Comparing Model\_A and Model\_D, it is hard to say which one is better. However, the AUC score calculated on the BBS dataset using features from \textit{conv4\_4} of Model\_D is noticeably better than that for Model\_A. This suggests that increasing the shape-bias of later layers of the CNN could potentially lead to better template matching. However, this results is not reflected by the results for the KTM dataset. One possible explanation is that the templates in the KTM dataset are smaller, in general, than those in the BBS dataset (if the template size is defined as the product of its width and height, then the mean template size of for the DIM dataset is 1603 whereas it is 3442 for the BBS dataset). 
Smaller templates tend to be less discriminative. The sub-sampling that occurs in later levels of the CNN results in templates that are even smaller and less disriminative. This may account for the worse performance of the later layers of each CNN when tested using the KTM dataset rather than the BBS dataset. It is also a confounding factor in attributing the better performance of the early layers to a reliance on texture information.
\subsection{Template matching using features from multiple convolutional layers}

We compared Model\_A and Model\_D by applying the DIM template matching algorithm to features extracted from multiple convolutional layers of each CNN. To combine feature maps with different sizes bilinear interpolation was used to make them the same size. If the template was small (height times width less than 4000) the feature maps from the later layer(s) were scaled to be the same size as those in the earlier layer(s). If the template was large, the feature maps from the earlier layer(s) were reduced in size to be the same size as those in the later layer(s). To maintain a balance between low and high level features, the dimension of the features maps form the latter layer(s) were reduced by PCA to the same number as those in the earlier layer.

Table~\ref{tab:table2} shows the AUC scores produced by DIM using features from two convolutional layers of Model\_A and Model\_D. All possible combinations of two layers were tested and the table shows only selected results with the best performance. Each cell of the table contains two AUC scores, the upper one is produced using Model\_A and the bottom is produced by Model\_D. The up and down-arrows indicate whether the AUC score of Model\_D is better or worse than that of Model\_A. 

\begin{table}[tb]
\centering
\caption{Partial AUC scores of DIM using features from two convolutional layers of Model\_A (upper value in each cell) and Model\_D (lower value in each cell).}
\scriptsize
\subtable[Evaluation on BBS dataset.]{
\scalebox{0.72}{\begin{tabular}{|c|l|l|l|l|l|l|}
\hline
\diagbox[width=5em,innerleftsep=0.2pt,innerrightsep=0.2pt,font=\tiny]{Layer}{AUC}{Layer}&\textit{conv3\_3}  &\textit{conv3\_4}&\textit{conv4\_1}&\textit{conv4\_2}&\textit{conv4\_3}&\textit{conv4\_4}\\\hline 
\multirow{2}{*}{\textit{conv1\_1}}      & 0.710            & 0.705           & 0.713           & 0.697           & 0.698          & 0.711             \\
                                        & 0.707$\downarrow$   & 0.714$\uparrow$ & 0.704$\downarrow$ & \textbf{0.718}$\uparrow$ & 0.710$\uparrow$& 0.708$\downarrow$ \\ \hline 
\multirow{2}{*}{\textit{conv1\_2}}      & 0.686             & 0.686          & 0.674           & 0.655           & 0.680          & 0.683             \\
                                        & 0.686   & 0.687$\uparrow$ & 0.707$\uparrow$ & 0.696$\uparrow$ & 0.690$\uparrow$& 0.710$\uparrow$   \\ \hline
\multirow{2}{*}{\textit{conv2\_1}}      & 0.658             & 0.670           & 0.664          & 0.653           & 0.662          & 0.667             \\
                                        & 0.659$\uparrow$ & 0.669$\downarrow$ & 0.665$\uparrow$ & 0.671$\uparrow$ & 0.683$\uparrow$ & 0.693$\uparrow$\\ \hline
\multirow{2}{*}{\textit{conv2\_2}}      & 0.659            & 0.661           & 0.653           & 0.641           & 0.659          & 0.663             \\
                                        & 0.665$\uparrow$   & 0.667$\uparrow$ & 0.676$\uparrow$ & 0.679$\uparrow$ & 0.676$\uparrow$& 0.682$\uparrow$   \\
\hline
\end{tabular}}
}
\scriptsize
\subtable[Evaluation on KTM dataset.]{
\scalebox{0.72}{\begin{tabular}{|c|l|l|l|l|l|l|}
\hline
\diagbox[width=5em,innerleftsep=0.2pt,innerrightsep=0.2pt,font=\tiny]{Layer}{AUC}{Layer}&\textit{conv3\_3}  &\textit{conv3\_4}&\textit{conv4\_1}&\textit{conv4\_2}&\textit{conv4\_3}&\textit{conv4\_4}\\\hline 
\multirow{2}{*}{\textit{conv1\_1}}      & 0.687             & 0.684             & 0.677           & 0.668           & 0.670           & 0.678             \\
                                        & 0.689$\uparrow$ & 0.691$\uparrow$ & 0.682$\uparrow$ & 0.695$\uparrow$ & 0.684$\uparrow$ & 0.687$\uparrow$ \\\hline  
\multirow{2}{*}{\textit{conv1\_2}}      & 0.687             & 0.689             & 0.682           & 0.685           & 0.675           & 0.682             \\
                                        & 0.680$\downarrow$   & 0.694$\uparrow$ & 0.695$\uparrow$ & \textbf{0.697}$\uparrow$ & 0.691$\uparrow$ & 0.690$\uparrow$\\ \hline 
\multirow{2}{*}{\textit{conv2\_1}}      & 0.634             & 0.633             & 0.647          & 0.645           & 0.655           & 0.639             \\
                                        & 0.642$\uparrow$   & 0.651$\uparrow$   & 0.665$\uparrow$ & 0.671$\uparrow$ & 0.668$\uparrow$ & 0.666$\uparrow$   \\\hline 
\multirow{2}{*}{\textit{conv2\_2}}      & 0.642             & 0.651             & 0.664           & 0.661           & 0.669           & 0.669             \\
                                        & 0.657$\uparrow$   & 0.664$\uparrow$ & 0.670$\uparrow$ & 0.673$\uparrow$ & 0.669 & 0.669 \\ \hline
\end{tabular}}
}
\label{tab:table2}
\end{table}

It can be seen from Table~\ref{tab:table2} that for the 24 layer combinations for which results are shown, 21 results for both BBS and KTM dataset are better for Model\_D than for Model\_A. Furthermore, the best result for each dataset (indicated in bold) is generated using the features from Model\_D. These results thus support the conclusion that more discriminative features can be obtained by slightly increasing the shape bias of the VGG19 model which increases the performance of template matching. 

\begin{table}[tb]
\centering
\caption{Best 10 results when using combinations of features from three convolutional layers of Model\_D. $C12^{41}_{44}$ means fusing the features from \textit{conv1\_2}, \textit{conv4\_1} and \textit{conv4\_4} for instance.} 
\scriptsize
\subtable[Evaluation on BBS dataset.]{
\scalebox{0.95}{\begin{tabular}{cccccccccccc}
\toprule[1pt]
Layers & $C12^{41}_{44}$ & $C11^{34}_{43}$ & $C11^{42}_{44}$ & $C12^{43}_{52}$ & $C11^{22}_{43}$ & $C11^{41}_{44}$ & $C11^{41}_{43}$& $C11^{34}_{42}$ & $C12^{43}_{44}$ & $C11^{34}_{44}$ \\
\\[-1em]
\hline 
\\[-1em]
AUC    & 0.728   & 0.727   & 0.724  & 0.724  & 0.723    & 0.722    & 0.720    & 0.720   & 0.720   & 0.720\\
\bottomrule[1pt]
\end{tabular}}
}
\subtable[Evaluation on KTM dataset.]{
\scriptsize
\scalebox{0.95}{\begin{tabular}{cccccccccccc}
\toprule[1pt]
Layers & $C11^{34}_{42}$ & $C12^{32}_{43}$ & $C11^{34}_{43}$ & $C12^{22}_{42}$ & $C12^{31}_{42}$ & $C12^{34}_{42}$ & $C12^{34}_{43}$& $C12^{31}_{43}$ & $C12^{33}_{43}$ & $C11^{33}_{42}$ \\
\\[-1em]
\hline 
\\[-1em]
AUC    & 0.711   & 0.709   & 0.708  & 0.706  & 0.706    & 0.705    & 0.705    & 0.705   & 0.705   & 0.704\\
\bottomrule[1pt]
\end{tabular}}
}
\label{tab:table3}
\end{table}

To determine if fusing features from more layers would further improve template matching performance, DIM was applied to all combinations of three layers from Model\_D. There are a total of 560 different combinations of three layers. It is impossible to show all these results in this paper, therefore the highest 10 AUC scores are shown in Table~\ref{tab:table3}. For both dataset, using three layers produced an improvement in the best AUC score (around 0.01) compared to when using two layers. 

\subsection{Comparison with other methods}

This section compares our results with those produced by other template matching methods in both color and deep feature-space. When evaluated on the BBS dataset, the deep features used by each template matching algorithm were the features from layers \textit{conv1\_2}, \textit{conv4\_1}, and \textit{conv4\_4} of Model\_D. When evaluated on the KTM dataset the deep features used as the input to each algorithm were those from layers \textit{conv1\_1}, \textit{conv3\_4} and \textit{conv4\_2} of Model\_D. BBS, CoTM and QATM have been tested on BBS data by their authors using different deep features, so we also compare our results to these earlier published results. 

The comparison results are summarised in Table~\ref{tab:table4}. All methods expect QATM and BBS produce improved results using the proposed deep features than when using color features. This is true for both datasets. Of the methods that have previously been applied to deep features the performances of two (NCC and QATM) are improved, and that of two others (BBS and CoTM) are made worse by using our method of defining the deep feature-space. Potentially, further improvements to the performance of these methods could be achieved by optimising the feature extraction method for the individual template matching algorithm, as has been done here for DIM.
However, it should be noted that simple metrics for comparing image patches such as NCC and ZNCC produce near state-of-the-art performance when applied to our proposed deep feature-space, outperforming much more complex methods of template matching such as BBS, CoTM, and QATM when these methods are applied to any of the tested feature-spaces, including that proposed by the authors of these algorithms. One known weakness of BBS is that it may fail when the template is very small compared to target image \cite{oron2017best}. This may explain the particularly poor results of this method when applied to the KTM dataset. 

DIM achieves the best results on both datasets when applied to deep features. DIM performs particularly well on the BBS dataset producing an AUC of 0.73 which, as far as we are aware, makes it the only method to have scored more than 0.7 on this dataset. The DIM algorithm also produces state-of-the-art performance on the KTM dataset when applied to deep features. When applied to color features, the results are good, although not as good as DDIS on the KTM dataset. This is because small templates in the KTM dataset sometime contain insufficient detail for the DIM algorithm to successfully distinguish the object. Using deep features enhances the discriminability of small templates sufficiently that the performance of DIM increases significantly. The results demonstrate that the proposed approach is effective at extracting distinguishable features which lead to robust and accurate template matching.
\begin{table}[tbh]
\caption{Quantitative comparison of the performance of different template matching algorithms using different input features.}
\centering
\scriptsize
\begin{tabular}{|c|c|c|c|c|c|}
\hline
\multirow{3}{*}{\diagbox[width=6em,innerleftsep=0.2pt,innerrightsep=0.2pt,height=3\line,font=\tiny]{Method}{AUC}{Feature}} & \multicolumn{3}{c|}{BBS dataset}                            & \multicolumn{2}{c|}{KTM dataset}    \\ \cline{2-6} 
                  & \multirow{2}{*}{Color} & \multirow{2}{*}{Deep}              & Deep              & \multirow{2}{*}{Color} & Deep       \\
                  &                        &                                    & (Proposed)        &                        & (Proposed) \\ \hline
SSD               & 0.46                   & -                                  & 0.54              & 0.42                   & 0.54       \\ \hline
NCC               & 0.48                   & 0.63 \cite{kim2017robust}         & 0.67              & 0.42                   & 0.67       \\ \hline
ZNCC              & 0.54                   & -                                  & 0.67              & 0.48                   & 0.67       \\ \hline
BBS               & 0.55                   & 0.60 \cite{oron2017best}            & 0.54              & 0.44                   & 0.55       \\ \hline
CoTM              & 0.54\footnotemark       & 0.67 \cite{kat2018matching}        & 0.64              & 0.51                   & 0.56       \\ \hline
DDIS              & 0.64                   & -                                  & 0.66              & \textbf{0.63}          & 0.68       \\ \hline
QATM              & -                      & 0.62\footnotemark    & 0.66              & -                      & 0.64       \\ \hline
DIM               & \textbf{0.69}          & -                                  &\textbf{0.73}      & 0.60                   & \textbf{0.71}       \\ \hline
\end{tabular}
\label{tab:table4}
\end{table}
\footnotetext[1]{We were unable to reproduce the result 0.62 reported in the paper \cite{kat2018matching} using code supplied by the authors of CoTM. Our different result is shown in the table.} 
\footnotetext[2]{The authors of QATM report an AUC score of 0.69 when this method is applied to the BBS dataset \cite{cheng2019qatm}. However, examining their source code we note that this result is produced by setting the size of the predicted bounding box equal in size to the width and height of the ground truth bounding box. Other methods are evaluated by setting the size of the predicted bounding box equal equal to the size of the template (i.e. without using knowledge of the ground truth that the algorithm is attempting to predict). We have re-tested QATM using the standard evaluation protocol and our result for the original version of QATM is 0.62. As QATM is designed to work specifically with a CNN it was not applied directly
to color images.}

\section{Conclusions}

Our results demonstrate that slightly increasing the shape bias of a CNN (by changing the method of training the network) produces more distinguishable features in which template matching can be achieved with greater accuracy. By running a large number of experiments we determined the best combination of convolutional features from our shape-biased VGG19 on which to perform template matching with the DIM algorithm. This same feature-space was shown to improve the performance of most other template matching algorithms as well. The DIM algorithm applied to our new feature-space produces state-of-art results on two benchmark datasets.

\subsection*{Acknowledgments}
The authors acknowledge use of the research computing facility at King’s College London, Rosalind (https://rosalind.kcl.ac.uk), and the Joint Academic Data science Endeavour (JADE) facility. This research was funded by China Scholarship Council.
 \bibliographystyle{splncs03_unsrt}
 \bibliography{references}

\begin{thebibliography}{10}
\providecommand{\url}[1]{\texttt{#1}}
\providecommand{\urlprefix}{URL }

\bibitem{bertinetto2016fully}
Bertinetto, L., Valmadre, J., Henriques, J.F., Vedaldi, A., Torr, P.H.:
  Fully-convolutional siamese networks for object tracking. In: European
  conference on computer vision. pp. 850--865. Springer (2016)

\bibitem{ma2018robust}
Ma, C., Huang, J.B., Yang, X., Yang, M.H.: Robust visual tracking via
  hierarchical convolutional features. IEEE transactions on pattern analysis
  and machine intelligence  (2018)

\bibitem{ahuja2013object}
Ahuja, K., Tuli, P.: Object recognition by template matching using correlations
  and phase angle method. International Journal of Advanced Research in
  Computer and Communication Engineering  2(3),  1368--1373 (2013)

\bibitem{dai2016r}
Dai, J., Li, Y., He, K., Sun, J.: R-fcn: Object detection via region-based
  fully convolutional networks. In: Advances in neural information processing
  systems. pp. 379--387 (2016)

\bibitem{scharstein2002taxonomy}
Scharstein, D., Szeliski, R.: A taxonomy and evaluation of dense two-frame
  stereo correspondence algorithms. International journal of computer vision
  47(1-3),  7--42 (2002)

\bibitem{chhatkuli2014stable}
Chhatkuli, A., Pizarro, D., Bartoli, A.: Stable template-based isometric 3d
  reconstruction in all imaging conditions by linear least-squares. In:
  Proceedings of the IEEE Conference on Computer Vision and Pattern
  Recognition. pp. 708--715 (2014)

\bibitem{chan2015pcanet}
Chan, T.H., Jia, K., Gao, S., Lu, J., Zeng, Z., Ma, Y.: {PCANet}: A simple deep
  learning baseline for image classification? IEEE transactions on image
  processing  24(12),  5017--5032 (2015)

\bibitem{wang2017residual}
Wang, F., Jiang, M., Qian, C., Yang, S., Li, C., Zhang, H., Wang, X., Tang, X.:
  Residual attention network for image classification. In: Proceedings of the
  IEEE Conference on Computer Vision and Pattern Recognition. pp. 3156--3164
  (2017)

\bibitem{liang2015recurrent}
Liang, M., Hu, X.: Recurrent convolutional neural network for object
  recognition. In: Proceedings of the IEEE conference on computer vision and
  pattern recognition. pp. 3367--3375 (2015)

\bibitem{wohlhart2015learning}
Wohlhart, P., Lepetit, V.: Learning descriptors for object recognition and 3d
  pose estimation. In: Proceedings of the IEEE Conference on Computer Vision
  and Pattern Recognition. pp. 3109--3118 (2015)

\bibitem{cheng2019qatm}
Cheng, J., Wu, Y., AbdAlmageed, W., Natarajan, P.: {QATM}: Quality-aware
  template matching for deep learning. In: Proceedings of the IEEE Conference
  on Computer Vision and Pattern Recognition. pp. 11553--11562 (2019)

\bibitem{kat2018matching}
Kat, R., Jevnisek, R., Avidan, S.: Matching pixels using co-occurrence
  statistics. In: Proceedings of the IEEE Conference on Computer Vision and
  Pattern Recognition. pp. 1751--1759 (2018)

\bibitem{kim2017robust}
Kim, J., Kim, J., Choi, S., Hasan, M.A., Kim, C.: Robust template matching
  using scale-adaptive deep convolutional features. In: 2017 Asia-Pacific
  Signal and Information Processing Association Annual Summit and Conference
  (APSIPA ASC). pp. 708--711. IEEE (2017)

\bibitem{oron2017best}
Oron, S., Dekel, T., Xue, T., Freeman, W.T., Avidan, S.: Best-buddies
  similarity—robust template matching using mutual nearest neighbors. IEEE
  transactions on pattern analysis and machine intelligence  40(8),  1799--1813
  (2017)

\bibitem{talmi2017template}
Talmi, I., Mechrez, R., Zelnik-Manor, L.: Template matching with deformable
  diversity similarity. In: Proceedings of the IEEE Conference on Computer
  Vision and Pattern Recognition. pp. 175--183 (2017)

\bibitem{kriegeskorte2015deep}
Kriegeskorte, N.: Deep neural networks: a new framework for modeling biological
  vision and brain information processing. Annual review of vision science  1,
  417--446 (2015)

\bibitem{geirhos2018imagenet}
Geirhos, R., Rubisch, P., Michaelis, C., Bethge, M., Wichmann, F.A., Brendel,
  W.: Imagenet-trained cnns are biased towards texture; increasing shape bias
  improves accuracy and robustness. arXiv:1811.12231  (2018)

\bibitem{spratling2019explaining}
Spratling, M.W.: Explaining away results in accurate and tolerant template
  matching. Pattern Recognition p. 107337 (2020)

\bibitem{kersten2004object}
Kersten, D., Mamassian, P., Yuille, A.: Object perception as bayesian
  inference. Annu. Rev. Psychol.  55,  271--304 (2004)

\bibitem{spratling2012unsupervised}
Spratling, M.W.: Unsupervised learning of generative and discriminative weights
  encoding elementary image components in a predictive coding model of cortical
  function. Neural computation  24(1),  60--103 (2012)

\bibitem{simonyan2014very}
Simonyan, K., Zisserman, A.: Very deep convolutional networks for large-scale
  image recognition. arXiv:1409.1556  (2014)

\bibitem{ren2015faster}
Ren, S., He, K., Girshick, R., Sun, J.: {Faster R-CNN}: Towards real-time
  object detection with region proposal networks. In: Advances in neural
  information processing systems. pp. 91--99 (2015)

\bibitem{huang2017arbitrary}
Huang, X., Belongie, S.: Arbitrary style transfer in real-time with adaptive
  instance normalization. In: Proceedings of the IEEE International Conference
  on Computer Vision. pp. 1501--1510 (2017)

\bibitem{wu2015object}
Wu, Y., Lim, J., Yang, M.H.: Object tracking benchmark. IEEE Transactions on
  Pattern Analysis and Machine Intelligence  37(9),  1834--1848 (2015)

\bibitem{liang2015encoding}
Liang, P., Blasch, E., Ling, H.: Encoding color information for visual
  tracking: Algorithms and benchmark. IEEE Transactions on Image Processing
  24(12),  5630--5644 (2015)

\end{thebibliography}

\end{document}